\documentclass[10pt,twocolumn,letterpaper]{article}

\usepackage{cvpr}

\usepackage{times}
\usepackage{epsfig}
\usepackage{graphicx}
\usepackage{amsmath}
\usepackage{amssymb}
\usepackage{multirow}
\usepackage{bm}
\usepackage[bottom]{footmisc}

\usepackage{adjustbox}
\usepackage{setspace}
\usepackage[normalem]{ulem}

\usepackage{booktabs}
\usepackage{subcaption}
\usepackage{parselines}
\usepackage{lineno}
\usepackage{setspace}
\usepackage{emptypage}
\usepackage{rotating}
\usepackage{booktabs}
\usepackage{makecell}
\usepackage{algorithm,tabularx}
\usepackage{csquotes}
\usepackage{xhfill}
\usepackage{xcolor}

\usepackage[pagebackref,breaklinks,colorlinks]{hyperref}

\usepackage[capitalize]{cleveref}
\crefname{section}{Sec.}{Secs.}
\Crefname{section}{Section}{Sections}
\Crefname{table}{Table}{Tables}
\crefname{table}{Tab.}{Tabs.}

\usepackage[accsupp]{axessibility} % Improves PDF readability for those with disabilities.

\begin{document}

%%%%%%%%% TITLE
\title{Towards Generalizable Morph Attack Detection with Consistency Regularization}

%\title{Generalizable Representation Learning for Morph Attack Detection.}

\author{Hossein Kashiani, Niloufar Alipour Talemi, Mohammad Saeed Ebrahimi Saadabadi,  Nasser M. Nasrabadi\\
West Virginia University\\
{\tt\small \{hk00014, na00027, me00018\}@mix.wvu.edu, nasser.nasrabadi@mail.wvu.edu
}}
%{\tt\small hk00014@mix.wvu.edu}


\maketitle
\thispagestyle{empty}

%%%%%%%%% ABSTRACT
\begin{abstract}

Though recent studies have made significant progress in morph attack detection by virtue of deep neural networks, they often fail to generalize well to unseen morph attacks. With numerous morph attacks emerging frequently, generalizable morph attack detection has gained significant attention. This paper focuses on enhancing the generalization capability of morph attack detection from the perspective of consistency regularization. Consistency regularization operates under the premise that generalizable morph attack detection should output consistent predictions irrespective of the possible variations that may occur in the input space. In this work, to reach this objective, two simple yet effective morph-wise augmentations are proposed to explore a wide space of realistic morph transformations in our consistency regularization. Then, the model is regularized to learn consistently at the logit as well as embedding levels across a wide range of morph-wise augmented images. The proposed consistency regularization aligns the abstraction in the hidden layers of our model across the morph attack images which are generated from diverse domains in the wild. Experimental results demonstrate the superior generalization and robustness performance of our proposed method compared to the state-of-the-art studies.

\end{abstract}

%%%%%%%%% BODY TEXT
\section{Introduction}
\label{sec:intro}

In face recognition systems, the face is tightly tied to an individual's identity \cite{saadabadi2023quality,zhao2019regularface}. By disrupting this particular link, morph attacks may pose a potential hazard \cite{venkatesh2020detecting,Sarkar2020,9841318}. As a result, morph attack detection plays an important role in face recognition systems \cite{damer2021pw}. Morph attacks take place when a single morphed image can be used to prove the existence of two or more different people. Morphed images are crafted by interpolating facial landmarks or the latent representations between two or more individuals.
\begin{figure}[t] 
\centering
    \includegraphics[width=0.49\textwidth]{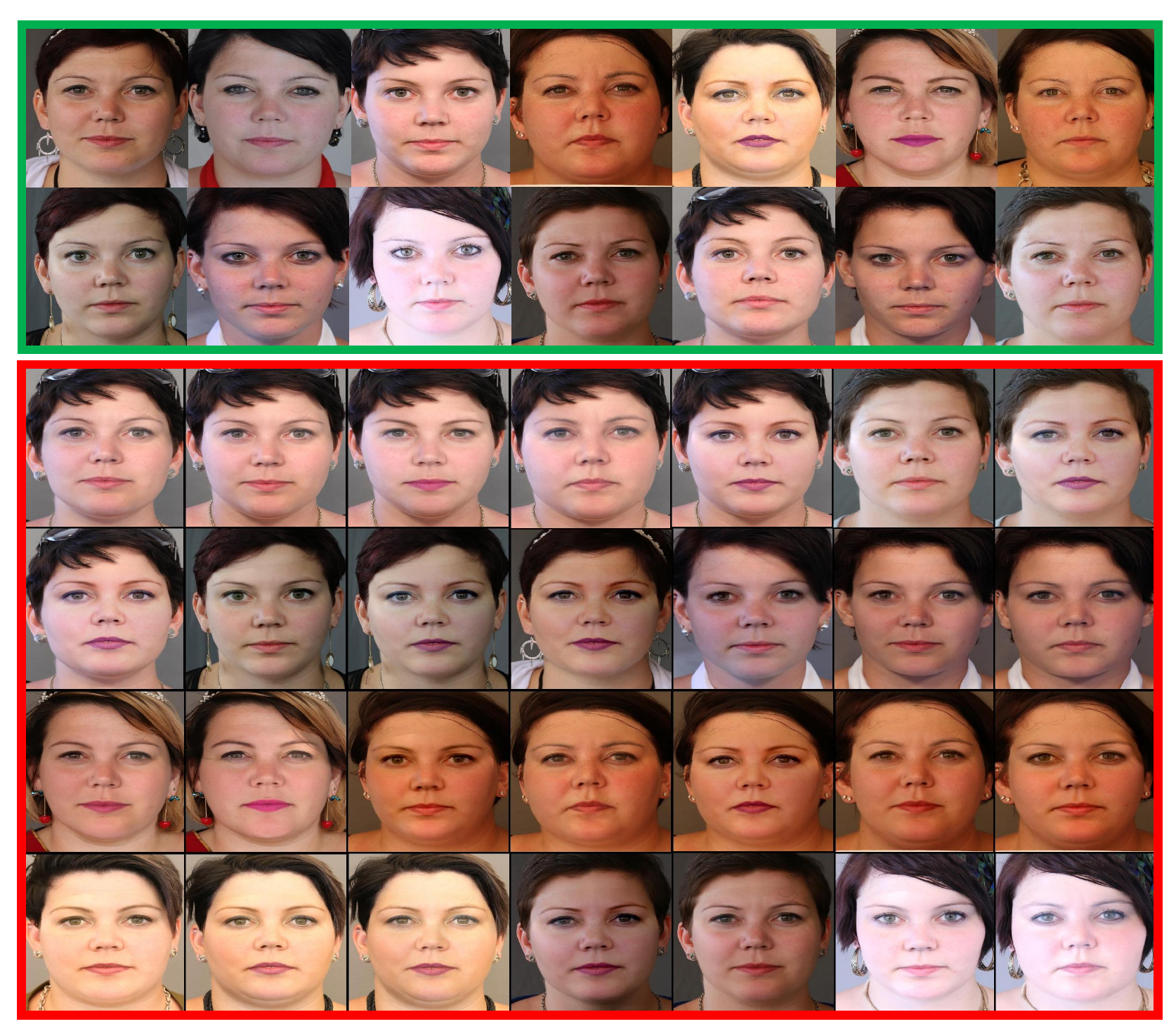}
    \caption{Illustration of the proposed SM augmentation for an identity sample in the Twin dataset \cite{TwinsIJCB}. The face images in the green and red bounding boxes correspond to the input bona fide and generated self-morphed images, respectively.}
    \label{fig1}
\end{figure}

Despite the success of previous morph attack detection studies in recent years \cite{neto2022orthomad,damer2021pw,raja2022towards}, their performance diminishes significantly on unseen morph attacks in real-world scenarios. A morph detection model should be generalizable to out-of-sample distributions, independent of the training distribution. Previous studies in morph detection \cite{ramachandra2022algorithmic,soleymani2021mutual} mainly overlook the domain shift issue in the real world and benchmark their methods on intra-domain scenarios. That is, the same methods are generally adopted in both training and test phases to create morph images. Moreover, generalization performance to post-processing operations such as print-scan conversion or JPEG compression are not taken into account in these works. As such, they obtain satisfactory performance on limited test sets, but struggle to retain its performance beyond their training regime. Using unlabeled target data, the domain adaptation (DA) methodology can be employed to mitigate the domain shift challenge \cite{kouw2019review}. However, in practical settings, we are not provided with the unlabeled target domain. To address this challenge in a more general setting, domain generalization methodology has attracted much attention recently \cite{zhou2022domain}. It aims to learn the domain invariant features  without accessing target domain data. \par
Recently, a few studies have been conducted to enhance the generalization of morph attack detection. 
For example, Damer et al. \cite{damer2019multi} embrace the observation that fusing several detectors which are trained on different types of morph attacks generalize better to new morph attacks. Also, in \cite{damer2021pw}, the pixel-wise supervision is incorporated into the binary classification to empower the morph detector with the generalization capability. However, these works \cite{damer2019multi,damer2021pw} typically rely on the diverse morph images generated from different morph attacks, which can be challenging to provide. Moreover, their generalization ability is restricted to certain situations far from real-world scenarios.

In this paper, we approach the generic morph attack detection differently by learning consistent feature representations. Our approach relies on the consistency regularization concept \cite{bachman2014learning,NIPS2016_30ef30b6,fan2022revisiting}, that a generic model should predict consistent results regardless of the plausible variations that the input images may undergo. Various factors may induce these variations, including brightness, lighting, image style, camera sensors, and the type of morphing attacks. Thus, a model with higher consistency under plausible common image corruptions such as noise, blur, contrast and compression corruptions is  expected to generalize better to new unexplored domains \cite{fan2022revisiting,zhu2021crossmatch,abuduweili2021adaptive}. Bearing these insights in mind, the consistent predictions are imposed on our model across a wide range of morph-wise augmented images. Building upon this concept, we can minimize the soft output class distributions between different morph attack variations. However, this minimization would not necessarily align the abstraction in the hidden layers of our model across morph attack images generated from different domains. To overcome this issue, we regularize our model to ensure consistent learning at both the logit and feature representation levels. For this objective, several regularization branches are first integrated into the intermediate layers of our model and the embedding levels are computed at these branches. Then, we jointly regularize feature representation as well as final soft output class distribution. To learn the domain-shared feature representation, adversarial feature learning \cite{li2018domain}
 is adopted among the morph-wise augmented images at different feature representation
levels. In this respect, a feature extractor competes with a domain discriminator to learn a domain-shared feature representation and the domain discriminator determines whether the input images come from the intact morph images or the augmented ones. An overview of our proposed architecture is illustrated in Figure \ref{fig_overal}.

To encourage our model to learn generic representation for morph attack detection, we incorporate cross-domain morph attacks into our consistency regularization. Note that blindly exposing a model to random image transformations does not necessarily enhance the cross-domain performance. Rather, it could hurt the inter-domain performance of our morph attack detection. With this consideration in mind, we propose two morph-wise augmentations, namely Inter-domain Style Mixup (ISM), and Self-morphing (SM) augmentations, to explore a wide space of realistic morph transformations in our consistency regularization. The ISM augmentation employs the photo-realistic style transfer \cite{yoo2019photorealistic} to synthesizes unseen morph attacks with new styles, while keeping the content of the input morph images unchanged. Also, the SM augmentation synthesize morph attacks with minimal visual artifacts using several instances of the same identity. The motivation of the proposed SM augmentation stems from the fact that in realistic morph attack scenarios, the visible morphing artifact are further post-processed and eliminated carefully. This process results in hardly recognizable but valuable morph images, which still contain imperceptible morphing artifacts. Our major contributions in this paper are
 
\begin{itemize}
    \item We regularize morph attack detection model to predict consistent results regardless of potential variations caused by diverse morph attacks, image quality, and  environmental situations.

    \item  We propose two morph-wise augmentations to explore a wide space of realistic morph attack transformations in our consistency regularization.

    \item We carry out extensive evaluations on several datasets to validate the generalization capability of our morph attack detection.

\end{itemize}

The remainder of this paper is structured as follows. Section \ref{sec1} provides a literature review of recent research in morph detection, domain generalization, and state-of-the-art data augmentation methods. Section \ref{sec3} describes our methodology in detail, including our proposed morph-wise data augmentation and the proposed consistency regularization. We provide comprehensive evaluations in section \ref{sec4} to assess the impact of synthetic morphs and proposed consistency regularization on the generalization performance. In this section, we also evaluate the generalization and robustness performance of our proposed morph detector compared with the state-of-the-art studies. Finally, Section \ref{sec5} concludes this paper.

\section{Related Work}\label{sec1}
\subsection{Morph Detection}

Morph attack detection studies \cite{damer2019multi,debiasi2019detection,9841318,raja2022towards,neto2022orthomad,damer2021pw,venkatesh2019morphed,kashiani2022robust} can be categorized into single and differential morph detection. Single morph detection attempts to distinguish the morphed image from the bona fide one. Differential morph detection, on the other hand, compares the potential morphed image with a second reliable image of the probe such as a live capture to  make its prediction. Recently, deep learning models have been widely used for morph detection. With the advent of deep learning, several morph detection studies have been carried out in recent years. Soleymani et al. \cite{soleymani2021mutual} train a disentangling network that produces disentangled representations for landmarks and facial appearance. They generate triplets of images, whereby each intermediate image takes the landmarks from one image and the appearance from the other image. To improve unknown re-digitized morph attacks detection, authors in \cite{damer2021pw} adjust pixel-wise supervision in the training to capture more informative morphing artifacts. Besides, they make a morphing dataset accessible to the public, which comprises digital and re-digitized morph attacks as well as bona fide images. Using a convolutional neural network, a de-morphing-based method is suggested in \cite{ortega2020border} to unravel the chip image and identify morphing presentation assaults in actual automated border control systems. Damer {\it et al.} \cite{damer2022privacy} create bona fide face images of non-existing people and develop a synthetic-based morph attack detection testset with StyleGAN2-ADA \cite{karras2020training}, whereby the legal and ethical difficulties associated with biometric data use can be mitigated. Equipped with SMDD dataset in \cite{damer2022privacy}, Ivanovska et al. \cite{ivanovska2022face} train Xception and HRNet networks to demonstrate the potential of synthetic morph data and justifies its importance for morph detection models across three limited morph datasets.

\subsection{Domain Generalization}

In supervised learning studies \cite{najafzadeh2023face,malakshan2023joint}, the training data is assumed to be from the same distribution as the test data . However, in most real-world scenarios with out-of-distribution (OOD) data, this assumption could be violated, and consequently, these algorithms suffer significant performance drops on an OOD data \cite{li2018domain,zhu2021crossmatch}. Domain generalization is intended to learn domain-invariant representations that are generalizable to an unseen domain based on labeled source domains \cite{zhou2022domain}. A number of studies in this area have been made with respect to data augmentation, domain alignment, ensemble learning, and self-Supervised learning. Regarding data augmentation, the studies \cite{yue2019domain,huang2021fsdr,wang2022feature,xu2021fourier} simulate domain shift through transferring the styles of source domain with external styles to learn domain-invariant representations.  In the domain alignment category, some researchers \cite{pan2018two,choi2021robustnet,zhou2020domain} work on normalization operations to eliminate information that aggravate domain shift issue.  Although domain generalization has achieved impressive results in image classification, object detection and semantic segmentation, little attention has been paid to the generalization capabilities of morph attack detection. In addition, the existing studies \cite{damer2019multi,damer2021pw} do not learn the invariant representations to the application-specific textural distortion. Therefore, we encourage our model to learn domain-invariant morphing attack feature representation which is found beneficial to mitigate domain shift challenge.

 \subsection{Data Augmentation}
Consistency-based methods rely on generating diverse yet reasonable augmentation of the input data. Data augmentation is one of the most effective solutions to improve model performance generalization without incurring computational cost in the inference time. Expanding the diversity of the training data with data augmentation can be regarded as a useful regularizer to mitigate overfitting issue \cite{jackson2019style,devries2017improved,zhang2018mixup}. It can also enhance the robustness of deep neural networks against input distribution shifts. Conventional data augmentations include simple label-invariant image transformations such as flipping, translation, jittering, and random cropping. As an example, CutOut randomly removes a square region in the input samples \cite{devries2017improved}. Recently, different studies on data augmentation have proposed to synthesize mixed samples and employ a sequence of image transformations. For instance, Mixup is the seminal study that linearly interpolates between two or more input samples to synthesize new samples \cite{zhang2018mixup}. Another group of studies such as style randomization \cite{jackson2019style} utilizes neural style transfer \cite{johnson2016perceptual} to modify the distribution of low-level features such as color and texture information in the training samples \cite{zhou2021domain,jackson2019style}. Based on the observations outlined in  \cite{geirhos2018imagenettrained}, they leverage Stylized ImageNet to mitigate the texture bias in deep neural networks and improve the generalization performance against distribution shifts.

\section{Methodology}\label{sec3}
\subsection{Problem Definition}
In the context of domain generalization, it is assumed that the data from a source domain $D_S$ is accessible to train our model. The ultimate objective is to train a model capable of performing as well as possible on data from unseen domains $D_T$, without requiring additional model updates based on target domain $D_T$. With no prior knowledge on $D_T$, we regularize our model to learn semantic consistency between several different landmark-based and GAN-based morphing attacks. These attacks include Print and Scan \cite{zhang2021mipgan}, StyleGAN2 \cite{Sarkar2020}, WebMorph \cite{Sarkar2020}, OpenCV \cite{Sarkar2020}, and FaceMorpher \cite{Sarkar2020} attacks. By doing so, we enforce our model to learn consistent representations in respect to different morphing attack artifacts rather than the domain-specific features relevant to the identity information.  Moreover, considering diversity and realism, we propose two morph-wise augmentations to synthesize novel morph domains and enrich the training source domain in the consistency regularization. In what follows, the proposed morph-wise augmentations are first explained. Then, the consistent regularisation learning is addressed.

\subsection{Morph-wise Augmentation}

\paragraph{Self-morphing Augmentation.} 

     The key idea in SM Augmentation is that hardly detectable morph attacks in reality with imperceptible morphing artifacts could enforce our model to learn more generalizable representations. As such, the SM Augmentation lies in the guidance of such high-quality morph attacks with minimal visible morphing artifacts, which are synthesized by two look-alike subjects. To promote the effectiveness of morph attack images, different instances of the same identity are employed in our morphing attacks. Formally, let $x_i$ represents an instance $i$, which belongs to the identity $x$. First, $x_i$ is randomly augmented with different image transformations, including Color, Gaussian noise, Blurring, Contrast, Brightness, Shear, Translate, and Compression operations. Then, the output self-morphed image would be calculated as follows:

\begin{equation}\label{eq.0}
 x_{ij}^{SM} =\Psi(\hat{x}_i,x_j),
\end{equation}
\noindent where $\hat{x}_i,x_j$ represent the augmented and pristine instances of identity $x$, and $x_{ij}^{SM}$ is the output self-morph image. Also, $\Psi$ denote the adopted morphing attacks, which include the StyleGAN, OpenCV, and FaceMorpher approaches. The SM Augmentation enriches the diversity of morph attacks so that our model can explore a large space of morphing artifacts and would not overfit to visible morphing artifacts. Figure \ref{fig1} represents an example of SM augmentation. 

\begin{figure}[t] 
\centering
    \includegraphics[width=0.5\textwidth]{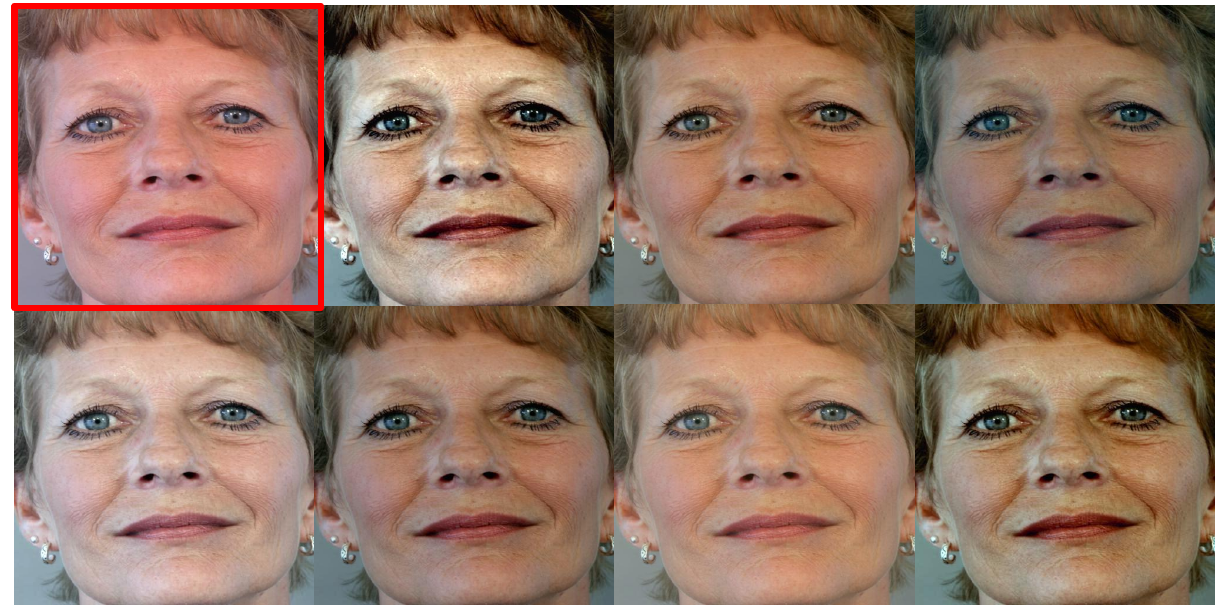}
    \caption{Illustration of the proposed ISM augmentation for an identity sample in the Twin dataset \cite{TwinsIJCB}. The face images in the red bounding box correspond to the input face image and the others indicate the augmented ones with the same class label.}
    \label{fig2}
\end{figure}

\begin{figure*}[t] 
\centering
    \includegraphics[width=0.98\textwidth]{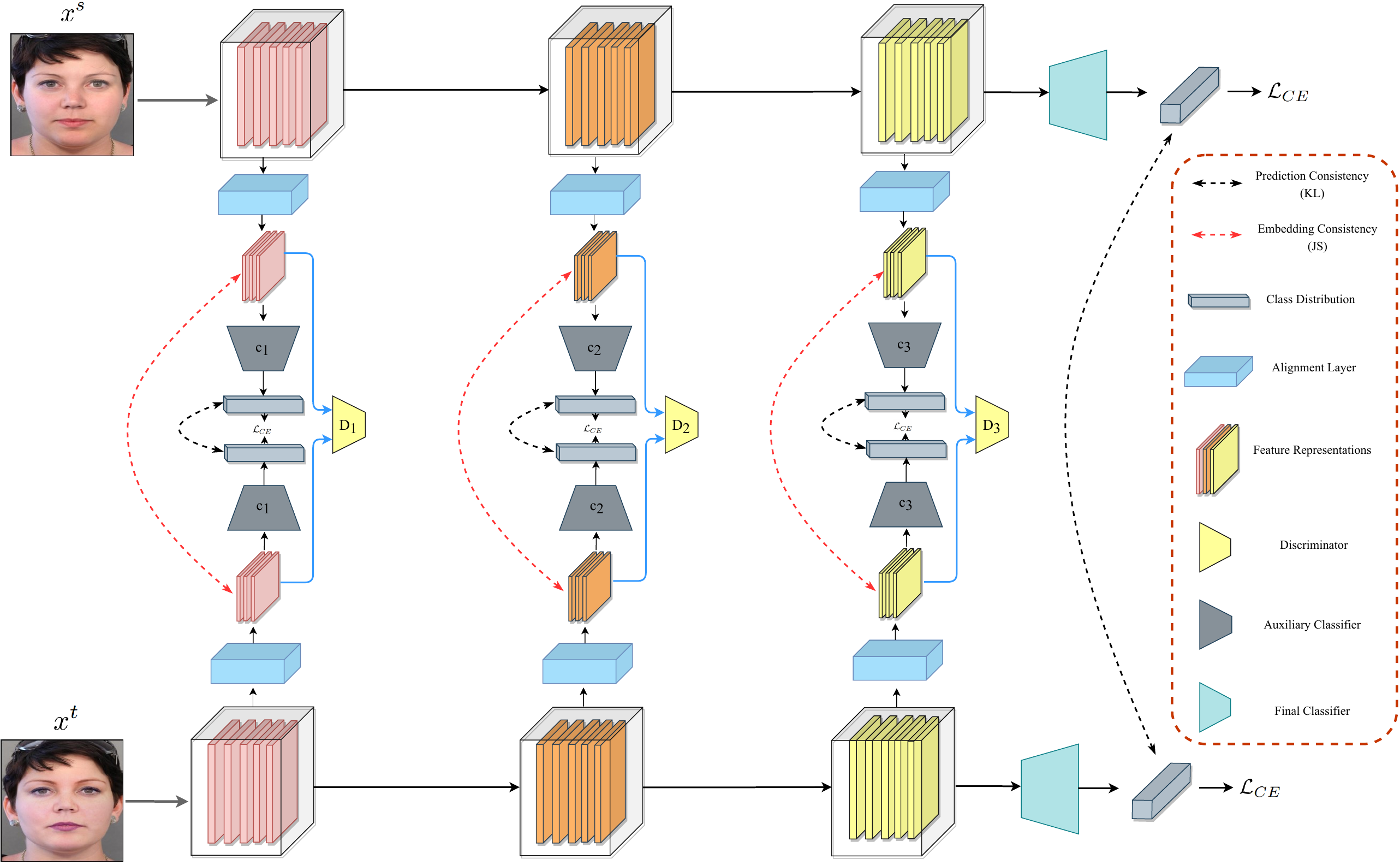}
    \caption{Illustration of the proposed architecture.}
    \label{fig_overal}
\end{figure*}

\paragraph{Inter-domain Style Mixup Augmentation.} 

ISM Augmentation aims to cover unconstrained scenarios that may occur in the real-world morph attacks in our consistency regularization. Equipped with the photorealistic style transfer method WCT2 \cite{yoo2019photorealistic}, ISM Augmentation manipulates the low-level style information in the source domain without compromising the high-level semantic information. This operation is formulated as follows:

\begin{equation}\label{eq.1}
 {x^{ISM}_{ij}} = \Omega(x^{co}_i, x^{st}_j),
\end{equation}

\noindent where $\Omega$ is the ISM transformation, the $x^{co}_i$, $x^{st}_i$, and $x^{ISM}_{ij}$ indicates the content image, the style image, and the output augmented image with a shared ground-truth label. Recall that the generated morph (or bona fide) images contain the content of the source morph (or bona fide) images and the style of the target morph (or bona fide) images. As such, the output generated synthetic images share identity information with the source images and low-level features with the target images. In a quest to find a dataset from a range of possibilities that can be used as the target domain, we opt for the FFHQ dataset \cite{karras2019style}. An example of ISM augmentation is shown in Figure \ref{fig2}.

\subsection{Consistency Regularization}

The baseline model is built from a feature extraction module with several levels (expressed as $K = \kappa _1 \circ \kappa_2 \circ ... \circ \kappa_K $) and a linear classifier $c_{f}$. Each level of the model is followed by a feature alignment module $\gamma_i$ and an auxiliary shallow classifier $c_i$. The feature alignment module $\gamma_i$ balances the feature dimension among different depths of the model so that the semantic abstractions would be regularizing throughout the network. Using the auxiliary classifier $\alpha_i$ and baseline model, the outputs at different levels can be obtained as below:

\begin{equation}\label{eq.2}
	\alpha_1 (x)= c_{1} \circ  \gamma_1 \circ \kappa_1(x), 
\end{equation}

\begin{equation}\label{eq.3}
	\alpha_i (x)= c_{i} \circ  \gamma_i \circ \kappa_i(x)... \circ \kappa_1(x), 
\end{equation}

\begin{equation}\label{eq.4}
	\mathrm{Baseline}(x)= c_{f} \circ \kappa_f(x) \circ \kappa_{f-1}(x) \circ ... \circ \kappa_1(x).
\end{equation}

While the logit outputs contain limited probabilistic information over classes, the feature representations at deep and shallow levels of the network encode richer information, which respectively capture the category-level semantic information and boundary content in the input images. Thus, an intuitive solution to setting up a powerful classifier is to incorporate a hierarchy of feature representations at several levels as $F_{cat} = \operatorname{Cat}([F_1,....,F_N])$. To do so, the feature representation $F_i$ is extracted at different levels by means of the auxiliary classifiers $c_i$ as $F_i=  \gamma_i \circ \kappa_i(x)... \circ \kappa_1(x)$, where $i \in {1,..,N-1}$, and $N-1$ is the number of levels in the baseline model. Afterwards, a linear classifier $c_{cat}$ operates on top of the concatenated features as given by:

\begin{equation}\label{eq.5}
	\alpha_{cat}(x)= c_{cat} \circ \operatorname{PWConv}\left(\operatorname{Cat}([F_1,....,F_N])\right),
\end{equation}

\noindent where $\operatorname{PWConv}$ denotes the point-wise convolution layer. Then, on a mixed set of raw and the morph-wise augmented images, the auxiliary classifiers ($c_{i\in 1,...,N-1,cat})$ and baseline backbone are trained using the standard cross-entropy (CE) loss function given as:

\begin{equation}\label{eq.6}
	{L_{cls}}= \sum_{i=1}^{N+1} \mathcal{L}_{CE}\left(\sigma(\alpha_i(x) ; \tau)),y\right),
\end{equation}

%5The associated softmax with temperature T is defined 

\noindent where $x$ and $y$ are the input sample and its groundtruth and $L_{CE}$ indicates the CE loss function. Also, $\alpha_N (x)$ and $\alpha_{N+1} (x)$ denote the final prediction of the baseline model and $c_{cat}$ is the classifier, respectively. Here, $\sigma(\alpha(x) ; \tau)$ denotes the Softmax operation with temperature $\tau$. An increase in $\tau>1$ leads to a softer probability distribution. The operation would be a normal Softmax if $\tau=1$.

\paragraph{Prediction-level Consistency Regularization}

To encourage the model to yield the same output distribution, we begin our consistency regularization with matching the class posterior distributions between predictions of the auxiliary classifiers $\alpha_i(x)$ for the source ${D_S}$ and augmented target domain ${D_T}$. Ideally, for even an unlabeled example, a robust model should produce consistent predictions no matter how it has been deformed and distorted. To achieve this, the Kullback–Leibler (KL) divergence minimization is applied as bellow:
\begin{multline}\label{eq.7}
	{L_{label}}= \sum_{i=1}^{N+1} D_{KL}\left({\sigma}(\alpha_i(x^s); \tau) ,{\sigma}(\alpha_{i}(x^t); \tau)\right),
\end{multline}
\noindent where $x_s \in{D_S}$, $x_t \in{D_T}$, and temperature scaling $\tau$ produces the soft output probability. Note that in the training process of morph class, ${D_T}$ includes the morph samples that are either augmented by the ISM and SM augmentations or are generated by a morph attack different from $x_s$. This regularizes a consistent posterior distribution with more comprehensive consistency whereby the classes with near-zero probabilities would not be simply discarded.  \paragraph{Embedding-level Consistency Regularization} We argue that a highly generalized model should behave consistently in feature representation space regardless of the styles and domains of the input images. Such representations encode the beneficial contents relevant to image intensity and spatial correlation. To fully meet this requirement, the feature representation $F_{i\in {{1,..,N-1}}}$ at different levels of the backbone model is extracted separately. Afterwards, the  feature representation $F_{i}$ are matched between the source ${D_S}$ and generated target domains ${D_T}$ by the Jensen-Shannon Divergence (JSD). We also integrate the discriminator ${Dsc_{i \in {1,..,N}}}$ (fed by the feature representation $F_i$) into our regularization framework to classify samples in the source ${D_S}$ domain from the generated target one ${D_T}$. These operations can be summarized as follows:

\begin{equation}\label{eq.10}
\begin{aligned}
	{L_{emb}}&= \sum_{i=1}^{N} D_{JS}\left({F_{i}}(x^{s}),{F_{i}}(x^{t})\right) + \eta  log \left(Dsc_{i}({F_{i}}(x^{s}))\right) \\&+ \eta  log \left(1- Dsc_{i}({F_{i}}(x^{t}))\right),
\end{aligned}
\end{equation}

\noindent where ${Dsc_{i \in {1,..,N}}}$ is trained with the CE loss function, and $D_{JS}$ refers to the JSD loss function. Compared with simple KL-divergence and Mean Square Error (MSE), JSD regularizes higher degree of consistency for feature representations and encourages more flexible optimization. Also, $\eta$ is the weight parameter that adjusts the importance of JSD regularization compared to CE optimization. Minimization in the Equation \ref{eq.10} realizes an adversarial process wherein the JSD regularization attempt to fool the discriminator ${D_{i \in {1,..,N}}}$ by learning indistinguishable feature representations.

\paragraph{Overall Loss}
Finally, the overall objective function in our training optimization can be summarized as follows:

\begin{equation}\label{eq.11}
L_{total} = {L_{cls}} + \mu{L_{label}} +\delta{L_{emb}} ,
\end{equation}

\noindent where $\delta$, and $\mu$ indicate the weighting parameters that balance the impact of different loss functions. In the inference step, the discriminators ${D_{i \in {1,..,N}}}$ and the auxiliary classifiers are detached and removed from the backbone model, thereby incurring no extra computational overhead compared to the baseline model.

\begin{table}

\center
\caption{Cross-morph evaluations of the proposed method with the state-of-the-art studies on FRGC datasets. The results are in terms of APCER1 (@BPCER=1\%), APCER5 (@BPCER=5\%), APCER (@BPCER10=10\%), EER, and AUC metrics. GAN refers to the StyleGAN2 \cite{Sarkar2020}.}

\resizebox{.48\textwidth}{!}{%
\begin{tabular}{cccccccc}
\toprule
\toprule
\addlinespace[1mm]

 &\multirow{1}{*}{{\makecell{Method}}} &  APCER1\%& APCER5\%&APCER10\% & EER &  AUC  \\
 \midrule
  % \multirow{7}{*}{}
\multirow{5}{*}{\rotatebox[origin=l]{90}{\makecell{MIPGAN}}}

& ConvNext \cite{liu2022convnet} &17.40 &3.07& 1.20& 16.33 & 99.17  \\
   \addlinespace[1mm]
 &Inception \cite{ivanovska2022face}&61.98 &36.68& 23.82& 17.26& 91.12 \\
 \addlinespace[1mm]

 &Residual \cite{9841318} &-&-& -& 6.67 & - \\
 \addlinespace[1mm]

&GRL&00.00 &00.00& 00.00 & 4.28& 99.99 \\
 \midrule
\multirow{4}{*}{\rotatebox[origin=l]{90}{\makecell{StyleGAN}}}

    & ConvNext \cite{liu2022convnet}
&44.60&14.52& 2.80& 7.65& 97.57  \\
   \addlinespace[1mm]
   
 &Inception \cite{ivanovska2022face}
 &50.60 &32.39& 25.56 & 17.26 &94.89  \\
 \addlinespace[1mm]

&GRL&00.00 &00.00& 00.00 & 00.00&100.00    \\

 \midrule
\multirow{4}{*}{\rotatebox[origin=l]{90}{\makecell{OpenCV}}}
   & ConvNext \cite{liu2022convnet}
&60.68 &29.66& 12.65 & 11.50 &95.27  \\
   \addlinespace[1mm]
   
 &Inception \cite{ivanovska2022face}
 &00.00 &00.00& 00.00 & 00.00&100.00  \\
 \addlinespace[1mm]

&GRL &00.00 &00.00& 00.00 & 00.00& 100.00 \\

\bottomrule
\end{tabular}%
}

\label{TableCross1}%
\end{table}%

\begin{table*}[!t]
\setlength{\tabcolsep}{0.6pt}

\center
\caption{ Cross-domain comparison of the proposed GRL with the state-of-the-art studies. The evaluations are in terms of the EER metric. }
\scriptsize 
\begin{tabular}{ccccccccccccccccccc}
\toprule
\toprule

%&& Methods &Target 1& Target 2 & Target 3 & Target 4& Target 5& Target 6& Target 7& Target 8& Target 9& Target 10& Target 11& Target 12& Target 13& Target 14& Target 15\\ \midrule
   && Methods  & 
   {\rotatebox[origin=c]{60}{\makecell{FRLL \\AMSL} }}& 
   {\rotatebox[origin=c]{60}{\makecell{FRLL\\ Webmorpher  }}}&
   {\rotatebox[origin=c]{60}{\makecell{FRLL \\OpenCV }}}&
   {\rotatebox[origin=c]{60}{\makecell{FRLL \\StyleGAN}}}&
   {\rotatebox[origin=c]{60}{\makecell{FRLL\\ FaceMorpher }}}&
   {\rotatebox[origin=c]{60}{\makecell{FERET\\ OpenCV}}}& 
   {\rotatebox[origin=c]{60}{\makecell{FERET\\ StyleGAN}}}&
   {\rotatebox[origin=c]{60}{\makecell{FERET\\ FaceMorpher}}}&
   {\rotatebox[origin=c]{60}{\makecell{FRGC \\OpenCV}}}& 
   {\rotatebox[origin=c]{60}{\makecell{FRGC \\StyleGAN}}}&
   {\rotatebox[origin=c]{60}{\makecell{FRGC \\FaceMorpher}}}&
   {\rotatebox[origin=c]{60}{\makecell{FRGC \\MIPGAN}}}&
   {\rotatebox[origin=c]{60}{\makecell{VISAPP17}}}&
   {\rotatebox[origin=c]{60}{\makecell{LMA-DRD }}}\\\midrule
 \addlinespace[1mm]
 \addlinespace[1mm]

  \multirow{13}{*}{\rotatebox[origin=l]{90}{\makecell{EER}}}

%&& Ensemble \cite{kashiani2022robust}&00.02&00.03
%&00.04&00.05&00.06&00.07&00.08&00.09&00.10 &00.11
%&00.12 &00.13&00.14&00.15&\\\addlinespace[1mm]

 &&SPL-MAD \cite{splmad}
 &12.09
 &15.72
 &5.78
 &12.92
 &4.67
 &30.21
 &28.95
 &25.76
 &19.54
 &15.57
 &18.42
 &-
 &
 &29.54
&\\
 %\addlinespace[1mm]
 
% &&Median Voting &99.98&99.98&92.94& 99.98  &96.33 &99.92 &95.36&94.65 &95.98  &99.84 &95.10&99.62&89.17&82.85 &\\
 \addlinespace[1mm]
 
 && MixFacenet \cite{ivanovska2022face}
 &15.18
 &12.35
 &4.39
 &8.99
 &3.87 
 &-
 &-
 &-
 &-
 &-
 &-
 &-
 &
 &23.72&\\
 \addlinespace[1mm]
 && Inception \cite{ivanovska2022face}
 &10.79 
 &9.86
 &5.38 
 &11.37
 &3.17  
 &-
 &-
 &-
 &-
 &-
 &-
 &-
 &
 &19.01&\\
 \addlinespace[1mm]
  && PW-MAD \cite{ivanovska2022face}
 &15.18
 &16.65 
 &2.42  
 &16.64 
 &2.20  
 &-
 &-
 &-
 &-
 &-
 &-
 &-
 &
 &20.39 &\\
 \addlinespace[1mm]

  && Hamza \cite{hamza2022generation}
 &-
 &-
 &-
 &-
 &-
 &13.5
 &-
 &11.5
 &-
 &-
 &-
 &-
 &-
 &-&\\ 
 \addlinespace[1mm]

  && Quality \cite{fu2022face}
 &7.91
 &7.13
 &5.41
 &7.04
 &3.60
 &12.29
 &13.99
 &10.80
 &24.48
 &14.32
 &24.17
 &-
 &
 &25.09&\\ 
 \addlinespace[1mm]

  && OrthoMAD \cite{neto2022orthomad}
 &14.80
 &15.23
 &0.73
 &6.54
 &0.98
 &-
 &-
 &-
 &-
 &-
 &-
 &-
 &-
 &-&\\ 

  \addlinespace[1mm]

  && Residuals (LMA) \cite{raja2022towards}
 &-
 &-
 &-
 &-
 &-
 &-
 &-
 &-
 &-
 &0.17
 &-
 &13.92
 &-
 &-&\\ 
 \addlinespace[1mm]
  && Mutual \cite{soleymani2021mutual}
 &3.11
 &-
 &-
 &-
 &-
 &-
 &-
 &-
 &-
 &-
 &-
 &-
 &4.69
 &-&\\ 
 \addlinespace[1mm]
  && Scale-Space Gradients \cite{9841318}
 &-
 &-
 &-
 &-
 &-
 &-
 &-
 &0.98 
 &-
 &-
 &- 
 &6.67 
 &-&\\

 \addlinespace[1mm]
 &&GRL
 &1.53
 &5.23
 &1.05
 &5.54
 &1.79
 &6.80
 &8.39
 &7.75
 &0.06
 &0.24
 &0.12
 & 0.40
 &0.00
 &13.09
 &\\

\iffalse
 &&GRL
 &1.53
 &5.2334
 &1.05
 &5.5483
 &1.79
 &6.8053
 &8.3970
 &7.7505
 &0.0608
 &0.2434
 &0.1217
 & 0.4016
 &0.00
 &13.09
 &\\
\fi
 \addlinespace[1mm]
\bottomrule
\bottomrule
\end{tabular}%

\label{TableCross3}%
\end{table*}%

\begin{table}

\center
\caption{Robustness evaluation of the proposed method against post-processing artifacts compared with the state-of-the-art studies on FRGC MIPGAN dataset. The results are in terms of APCER1 (@BPCER=1\%), APCER5 (@BPCER=5\%), APCER (@BPCER10=10\%), EER, and AUC metrics. PS and JPG refer to the MIPGAN-PS and MIPGAN-JPG test sets.}

\setlength{\tabcolsep}{3pt}

\scriptsize 

\begin{tabular}{cccccccc}
\addlinespace[3mm]

\toprule
\toprule
\addlinespace[1mm]

 &\multirow{1}{*}{{\makecell{Method}}} &  APCER1\%& APCER5\%&APCER10\% & EER &  AUC  \\
 \midrule
  % \multirow{7}{*}{}
\multirow{5}{*}{\rotatebox[origin=l]{90}{\makecell{PS}}}

   & ConvNext \cite{liu2022convnet}
&95.71&75.90& 60.50 & 31.19& 76.11 \\
   \addlinespace[1mm]

 &Inception \cite{ivanovska2022face}
&86.88 &71.48& 54.48 & 27.40 & 77.73 \\
 \addlinespace[1mm]

 &Residual \cite{9841318} &-&-& -& 9.63 & - \\
 \addlinespace[1mm]

&GRL  &67.60&19.41& 9.63 & 10.84&95.28  \\
 \midrule
\multirow{4}{*}{\rotatebox[origin=l]{90}{\makecell{JPG -112}}}
   &ConvNext-112 \cite{liu2022convnet}
&95.71 &75.90& 60.50&31.19&94.73    \\
   \addlinespace[1mm]

    &Inception-112 \cite{ivanovska2022face}
&76.97&49.13& 37.35 & 21.48&  86.65\\
 \addlinespace[1mm]
&GRL-112 &39.75 &6.024& 1.74&  5.67 & 98.08\\
 \addlinespace[1mm]

  \midrule
\multirow{4}{*}{\rotatebox[origin=l]{90}{\makecell{JPG -64}}}
      &ConvNext-64 \cite{liu2022convnet}
&61.98& 36.68 &23.82&17.77&91.36  \\
   \addlinespace[1mm]

 &Inception-64 \cite{ivanovska2022face}
&91.83&75.23& 57.69 & 27.65 & 78.21\\
 \addlinespace[1mm]
&GRL-64 &37.75&16.46& 9.10&  9.50 & 96.63\\
 \addlinespace[1mm]
  \midrule

&GRL-32 &65.59& 42.83&27.84  &16.33& 90.75\\

 \bottomrule
\bottomrule
\end{tabular}%

\label{TableCross2}%
\end{table}%

\section{Experiments}\label{sec4}

This section provides the explanation about the test and train datasets, the implementation details and evaluation protocols. Also, a comparative evaluation is performed to demonstrate that the proposed method 
  with generalizable representation learning (called GRL) significantly outperforms its competitors in respect to the generalization and accuracy performance.

\subsection{Evaluation Settings}\label{evalsec}

To fully assess the generalization capability of our morph attack detection, the experimental evaluations are carried out in two settings. In the first setting, we study the generalization performance of our method from one morph attack to the unseen attacks. More specifically, the bona fide images remain unchanged, yet the domain discrepancy exists in the morph attack and morph artifacts. In this setting, the FRGC dataset \cite{phillips2005overview} is adopted as the training data and the morph attacks are generated via the FaceMorpher method \cite{Sarkar2020}. Also, the target domain belongs to the FRGC morph faces which are created by the other morph attacks such as StyleGAN2, MIPGAN, and OpenCV approaches. When employed in an identification document, images may undergo various post-processing operations such as JPEG compression, resizing, and print-scan transformations, leading to the new types of artifacts in addition to the morph ones. In this regard, to benchmark the robustness of our method against these artifacts, we craft the printed-and-scanned MIPGAN (MIPGAN-PS) and JPEG compressed (MIPGAN-JPG) test sets using the FRGC images.

In the second setting, to perform morph detection on a cross-domain dataset, we employ the Twins morph dataset \cite{TwinsIJCB} as the training set. The Twins dataset is composed of 9,052 bona fide and 12,991 morphed images. To generate high-quality morphs in this dataset, identical twin pairs are selected as the contributing subjects in the landmark-based and Generative Adversarial Network (GAN)-based morphing methodologies. The FaceMorpher library \cite{Sarkar2020} and the pre-trained StyleGAN2 model \cite{karras2020training} are utilized as the former and latter methodologies, respectively. To corroborate the effectiveness of morph detection method over morph images with different distributions compared to the training set, FRGC \cite{phillips2005overview}, AMSL \cite{neubert2018extended}, FERET \cite{FERET}, VISAPP17 \cite{neubert2018extended}, and FRLL \cite{debruine2017face,Sarkar2020} datasets are targeted in our experimental evaluations. To fully study the generalization capability of our morph detection model, we also benchmark our model against a wide range of landmark-based and GAN-based morphing attacks. These attacks consists of Print and Scan \cite{zhang2021mipgan}, StyleGAN2 \cite{Sarkar2020}, WebMorph \cite{Sarkar2020}, OpenCV \cite{Sarkar2020}, and FaceMorpher \cite{Sarkar2020} attacks. At last, we also conduct ablation experiments to validate the importance of SM and ISM augmentations in our proposed consistency regularization.
\paragraph{Evaluation Metrics.} To gauge the performance of our morph attack detection, the Attack Presentation Classification Error Rate (APCER) is adopted. This metric computes the ratio of morph attacks which are incorrectly classified as bona fide. Also, to gain a comprehensive performance of the morph attack detection, the Area-Under-the-Curve (AUC) and Detection Equal Error Rate (D-EER) are computed. It is worth noting that the D-EER reports the classification error where APCER is equal to BPCER.
\paragraph{Implementation Detail.}
To pre-process the training data, the MTCNN model \cite{zhang2016joint} is utilized to detect and align face images. Then, the captured faces are re-scaled to  $512\times 512$ resolution. The ConvNext network \cite{liu2022convnet} is also selected as the backbone model. In order to train our backbone model, the Stochastic
Gradient Descent (SGD) with momentum $0.9$ is employed. The initial learning rate, batch size, and the total number of epochs are set to $1e-4$, $64$, and $50$, respectively. The hyperparameters used in Equations \ref{eq.6}, \ref{eq.10}, and \ref{eq.11} are set to $\tau =0.1 $, $\eta =0.1$, $\mu =0.05$, and $\delta =0.1$.

\subsection{Results on Unseen Morph Attacks}
In this evaluation setting, we measure how well the learned representations in the domain-specific morph artifacts may transfer to other types of morph artifacts. From Table \ref{TableCross1}, we can observe that the proposed method achieves much higher generalization performance under all evaluation metrics over the baseline model (ConvNext) and other morph detection models when tested on the MIPGAN, StyleGAN2, and OpenCV morph attacks. For instance, compared to the baseline model, our approach improves the APCER1\% on MIPGAN, StyleGAN2 and OpenCV morph attacks from 17.4\%, 44.60\%, and 60.68\% to 0\%. The reason behind such a pronounced generalization is that the proposed consistency regularising enforces the model to learn domain-agnostic feature representations. These results demonstrate the effectiveness of the proposed GRL to greatly benefit the out-of-distribution generalization of the morph attack detection.

\subsection{Results on Unseen Post-processing Artifacts}

The objective of this experiment is to benchmark how robust the GRL is against new types of artifacts induced by post-processing operations. 
The vulnerability assessments of our method against Print/Scan and JPEG compression operations are presented in Table \ref{TableCross2}. The FRGC dataset is the target test set wherein the morph attacks are generated by the MIPGAN approach. It is evident from the results in this analysis that the proposed consistency regularization equipped with the morph-wise augmentations can potentially gain considerable robustness against unseen post-processing artifacts in morph attack generation compared with vanilla morph attack detection studies. In Print/Scan and JPEG compressed morph images, our morph detection model significantly outperforms the other studies. It is worth highlighting that the performance of the proposed GRL against compressed images with resolution $32\times32$ surpasses the baseline model against compressed images with resolution $128\times128$.

\subsection{Results on Unseen Datasets}
As discussed previously, in the second setting, we explore the generalization capability of the proposed GRL to a wide range of unseen bona fide images, the domain of which are far away from our training set. In addition, since the distribution of morph images is also of great importance, unseen morph attacks are also integrated into the test data as well. As reported in Table \ref{TableCross3}, our GRL remarkably outperforms the state-of-the-art methods. In our evaluation, the best results of other studies are reported for comparison evaluations. For instance, in OrthoMAD, the best performance with the SMDD training set are reported. This is also the case for other studies. An in-depth analysis in Table \ref{TableCross3} reveals that the superiority of GRL over the state-of-the-art studies such as Quality \cite{fu2022face} and SPL-MAD \cite{splmad} is more noticeable on FRGC dataset than the others. The results validate that the state-of-the-art studies significantly lag behind the proposed GRL in both out-of-distribution and in-distribution morph attack detection. Note that on some test sets such as FRLL, the proposed GRL achieves better results on in-distribution morph attacks compared to the out-of-distribution morph attacks; however, this gap cannot be observe in other test sets such as FRGC. In short, taking these results into account, we can substantiate the generalization capability of GRL in out-of-distribution morph attack detection while retaining its high in-distribution performance for morph attack detection.

\subsection{Ablation Study}
To determine the contribution of SM, and ISM augmentations and also the embedding- and prediction-level consistency regularizations in the proposed morph detection, we eliminate them from the proposed GRL and follow the second setting of our training as mentioned in subsection \ref{evalsec}. Then, the trained degraded versions of GRL are assessed individually on the MIPGAN test set. The results in Table \ref{TableAblation} verify that each one of the proposed components, namely SM, and ISM augmentations,  ${L_{label}}$ and ${L_{emb}}$ consistency regularizations, plays an important role in the generalization and robustness performance of the proposed morph detection. Interestingly, SM augmentation shows higher gains in AUC and EER metrics than ISM augmentation. ISM augmentation synthesizes high-quality morphs images with minimal visual artifacts using a wide range of instances of the same identity, which is vital for morph detection. Furthermore, ${L_{emb}}$ consistency regularization leads to more significant improvements in AUC and EER metrics compared to ${L_{label}}$ consistency regularization. Moreover, we perform an additional ablation study to assess the impact of weight hyperparameters and embedding levels on generalization performance (see Table \ref{Table_last}). Reducing the weight of embedding-level consistency regularization ($\delta$) compared to prediction-level consistency regularization ($\mu$) resulted in decreased performance on the MIPGAN test set. Additionally, a significant drop of $1.34\%$ occurred when the number of embedding levels was reduced from three to one.

\begin{table}
\setlength{\tabcolsep}{6pt}
\scriptsize 
\center

\caption{Ablation evaluations of the proposed method on MIPGAN test set. The results are in terms of EER, and AUC metrics.}

\begin{tabular}{ccccccc}
\addlinespace[3mm]

\toprule
\toprule
\addlinespace[1mm]
 Metric&Baseline &  + SM &+ ISM &+ ${L_{label}}$ &+ ${L_{emb}}$& GRL\\
 \midrule
 EER   &11.24 &4.21& 9.47&6.74&1.79&0.4  \\
   \addlinespace[1mm]
 AUC   &96.03&99.18& 97.45&97.91&99.50&99.95 \\
 \bottomrule
\bottomrule
\end{tabular}%
\label{TableAblation}
\end{table}%

\begin{table}
\setlength{\tabcolsep}{6pt}
\scriptsize 
\center
\caption{Ablation studies on the weight parameters, number of embeddings. The results are in terms of EER, and AUC metrics.}
\begin{tabular}{cccccc}
\addlinespace[3mm]

\toprule
\toprule

 $\mu$  & $\delta$ &$N_{level}$   & SM + ISM + ${L_{label}}$ + ${L_{emb}}$& AUC&EER \\
 \midrule

 $0.05$ & $0.1$&3&$\checkmark$& ${99.95}$& 0.4\\
 $0.5$ & $0.1$&3&$\checkmark$&${99.58}$ &1.65\\
 $0.1$ & $0.1$&3&$\checkmark$&${99.83}$& 0.73\\
 $0.05$ & $0.1$&1&$\checkmark$&${98.64}$ &5.69\\
  \bottomrule
\bottomrule

\end{tabular}
\label{Table_last}
\end{table}

\section{Conclusions}\label{sec5}
In this paper, we present a morph attack detection with strong generalization ability to different morph attacks. To make our detector generalize better to unseen face morph attacks, we propose the ISM and SM morph-wise augmentations to explore a wide space of realistic morph attack artifacts in our consistency regularization. The ISM augmentation synthesizes unseen morph attacks with new styles, whilst preserving the content of the input morph images. Moreover, the SM augmentation generates realistic morph attacks with imperceptible visual morph artifacts. To improve the generalization performance of our detector against unseen face morph attacks, we encourage our model to predict consistent output regardless of the input variations simulated for different domains. To this end, we regularize our model to learn consistently at the logit and feature representation levels. Experimental results on several datasets demonstrate the generalization ability of our proposed model while keeping high in-domain performance.\\
\textbf{Acknowledgements.} This material is based upon a work supported by the Center for Identification Technology Research and the National Science Foundation under Grant \#1650474.

{\small
\bibliographystyle{ieee}
\bibliography{paper.bib}
}

\end{document}